# Human Machine Co-Creation. A Complementary Cognitive Approach to Creative Character Design Process Using GANs

Mohammad Lataifeh*[a], Xavier A Carrasco[a], Ashraf M Elnagar[a], Naveed Ahmed[a], Imran Junejo[b]

[a]*Department of Computer Science, University of Sharjah, United Arab Emirates*
*Corresponding author: mlataifeh@sharjah.ac.ae*
[b]*Advanced Micro Devices, AMD, Markham, ON, Canada*

**Abstract**

Recent advances in Generative Adversarial Networks (GANs) applications continue to attract the attention of researchers in different fields. In such a framework, two neural networks compete adversely to generate new visual contents indistinguishable from the original dataset. The objective of this research is to create a complementary co-design process between humans and machines to augment character designers' abilities in visualizing and creating new characters for multimedia projects such as games and animation. Driven by design cognitive scaffolding, the proposed approach aims to inform the process of *perceiving*, *knowing*, and *making*. The machine generated concepts are used as a launching platform for character designers to conceptualize new characters. A labelled dataset of $22,000$ characters was developed for this work and deployed using different GANs to evaluate the most suited for the context, followed by mixed methods evaluation for the machine output and human derivations. The discussed results substantiate the value of the proposed co-creation framework and elucidate how the generated concepts are used as cognitive substances that interact with designers' competencies in a versatile manner to influence the creative processes of conceptualizing novel characters.

# 1. Introduction.

Despite the state-of-the-art performance of machine learning algorithms, humans have always performed exceptionally better than any machine in different areas (Funke et al., 2019), and we believe this will continue to be the case specifically while evaluating the perceptual value of a designed concept. Humans, however, are trapped within the boundaries of their cognitive abilities. Particularly so, while challenged by the continuous demand of creating new design concepts of characters deployed in a wide array of multimedia projects. GANs provide a possible solution for such cognitive limits, by generating an abundance of visual stimuli of unfinished design concepts that become a stepping-stone for character designers to build on. Hence, creating the ultimate complementary co-design process between humans and machines. Indeed, many of the recently hailed applications for generative models are merely computed syntheses of human creative work from platforms such as ArtStation (Weatherbed, 2022), where designers protested the platform decision allowing their creation to be scrapped without permission as training materials (Baio, 2022; Growcoot, 2023), particularly as the generated outputs are often framed as cheaper, faster, and superior to that of humans. The debated views on this are far from being settled, and while we see the value provided by different generative models, we believe our proposed framework concerned with the process, not the product, provides a blueprint for moving forward.

Other than a visual novelty, innovating a new design concept for a character must also fit a specific narrative or context. Therefore, despite the gallant strides of success in a wide range of implementations, creating a realistic result of life-like visual scenes and characters (Karras et al., 2019) does not work well for this creative domain. Ideally, machines' generative process must be interrupted by humans (McFarlane & Latorella, 2002) as the output product is not the aim here, but rather the interaction of the visual stimuli with the cognitive process of designers to augment(Liao et al., 2020) and visibly (Barnard & May, 1999) support such a creative process. Designers exhibit different levels of expertise that set them apart from novices, especially in terms of their awareness of the design process complexity, moving from a design brief to sketches, shapes, volumes, themes, colors, and textures of a new concept. Design expertise is already an established reality for the design and engineering community (Dorst & Reymen, 2004; Dreyfus & Dreyfus, 1986, 2005). However, the constant high demands of new character concepts may induce different limitations of cognitive abilities referred to as *designers' burnout*, a recently classified occupational phenomenon by the World Health Organization(WHO, 2019). The same issue had already been raised decades earlier while discussing the concept of flow and creativity (Csikszentmihalyi, 1988).

Successful designers, reaching a high level of exhaustion in their creative production, may exhibit some limitations in their work, falling within analogical or stylistic similarities of previously presented concepts. As thinking is channeled by both context and goals, along with their associations in memories, character designers may fall into what psychologists call *mental set* or *fixation* (Jansson & Smith, 1991; Viswanathan & Linsey, 2012). As a result, rich knowledge of situations, goals, and materials, coupled with the internalized mental processes of design, are overshadowed by unconscious adherence to specific mental actions, and often, narrow memory recalls that affect the output of conceptual design. Such tacit knowledge inherits previous constraints which invariably influence the newly generated form or shape of a character, to an extent that a creative solution may no longer fulfill the need for novel concepts.

Furthermore, designers' perception stimulated by a visual proposition is usually sharper and more stable compared to mental images formed by remembered representations of features, objects, structures, and semantics (Fish & Scrivener, 1990; Goldschmidt, 1991). Designers, therefore, employ different strategies to step into fresh ground to inform their creative process while creating novel outputs. As such, the generated visuals are proposed as a non-verbal depiction of a design brief, or the starting point for designers to synthesize, formulate, evaluate, and reflect (Lawson, 1980; Nelson & Stolterman, 2003). Therefore, this work aims to bridge such interactive processes between humans and machines as one novel framework for catalyzing this endeavor.

The remainder of the paper is organized as follows: research objectives and contributions are listed in Section 2. Related work is presented in Section 3. Section 4 describes the datasets and models used. Followed by the details of the experiments in Section 5. Ending with the discussion and conclusion in Sections 6 and 7.

## 2. Research Objectives.

While the majority of recent GANs and other generative models are focused on realism within the generated outcome, our main objective is to provide a near-optimal visual result to stimulate and inform the creative process while conceptualizing new characters, proposed as a novel co-design process integrating machines intelligence with human creativity.

Within this unique context, the initial challenge to be addressed is to articulate a combination of objective and subjective measures for evaluating the performance and limitations of the generated outcomes. Indeed, there has been no consensus on one particular metric to evaluate GANs due to the diversity of applications. Further details were discussed in (Borji, 2019), which include several quantitative and qualitative measures, the adoption of which would certainly be contextually dependent.

Nonetheless, to stimulate the creativity and aptitudes of concept designers to create novel characters, throughout this work we explore different GAN architectures and their performance as tested on our new dataset called **Characters_silhouettes**. To measure the performance, quality, and usefulness of the obtained results, two methods were employed. Fréchet Inception Distance (FID) (Heusel et al., 2017) is used first to assess the quality of the generated images compared to ground truth. However, to address the creative usability of the generated results, an adaptation of Heuristic Review and Cognitive Walk-through (Molich & Nielsen, 1990) were used with the creative work produced by five character designers who were invited to use the proposed framework to create new concepts. Participants were also requested to maintain a commitment to a think-aloud protocol (Payne, 1994) to help externalize the cognitive processes for the visual information from the selected silhouette to the final work.

To the best of our knowledge, this is the first work of its kind that integrates GANs for such a creative context. Hence, the novelty of this work can be summarized as follow:

- A novel framework for character design conceptualization integrating machine intelligence for augmenting human creativity.

- Elucidating the interactions between GANs output and creative design

process at different levels of designers' expertise.

- Creating a web-based utility to interact with a modified adaptation to the state-of-the-art models for the non-technical creative audience to generate concepts with particular attributes.

- A new dataset containing 22, 000 labelled characters is shared publicly [1] to open doors for diverse research on visual and cognitive stimulation, particularly for creative applications.

To understand how this work is carried out, the following section will explore related work in the literature and how these generative networks have evolved to the possibilities they currently offer.

## 3. Related Work.

Since its inception, the first GAN (Goodfellow et al., 2014) caught the interest of many researchers for the new algorithmic directions offered, where two networks compete adversely to generate new images indistinguishable from the original dataset. Most of the initial improvements were related to the techniques and type of networks used in training, notably integrating the Convolutional Neural Networks (LeCun et al., 1998) referred to as Deep Convolutional GAN (DCGAN). Despite its distinguished output, DCGANs had three main limitations: the impossibility to deal with high-resolution images, mode collapse (Theis et al., 2015), and somewhat reliance on conditioned output instead of a completely random image. Addressing the main issue of mode collapse, works like Wasserstein GAN (WGAN) (Arjovsky et al., 2017), WGAN with Gradient Penalty (WGAN-GP) (Gulrajani et al., 2017), GAN-QP (Su, 2018) and SparseGAN (Mahdizadehaghdam et al., 2019) proposed changes in the loss function and the training process, providing an effective way to avoid mode collapse, despite increasing the training time. Further details on GANs and their different versions are covered in (Creswell et al., 2018; Hong et al., 2019)

On the other hand, the lack of datasets containing high-resolution images (higher than 512 × 512) may have slowed down improvements in the output's quality, and many researchers only focused on improving the results obtained for well-known datasets like CIFAR10 (Krizhevsky et al., 2006), LSUN (F. Yu et al., 2015), and ImageNET (Deng et al., 2009). The first successful attempt to work with 512×512 was BIGGAN, BIGGAN-deep (Brock et al., 2019) and BiBigGAN (Donahue & Simonyan, 2019) trained on ImageNET and JFT-300M (Hinton et al., 2015) datasets, which demonstrated significant improvements compared to earlier work, albeit being computationally demanding using at least 4 GPUs. The introduction of the CelebA-HQ dataset in the Progressive Growing of GANs (Karras et al., 2017) and further implementation of this technique in StyleGAN (Karras et al., 2019) became a turning point, providing a new method to deal with high-resolution images in GANs. Further modifications of this process were presented subsequently in StyleGAN2 (Karras, Laine, et al., 2020) and StyelGAN2-ada (Karras, Aittala, et al., 2020).

---

[1]The dataset includes 800K additional silhouettes generated to calculate FID scores can be accessed via https://data.mendeley.com/datasets/sdwbf4xrwz/3

Another challenge for the original models of GANs was about conditioning the generated images and allowing the possibility to choose the type of images to be generated by the model. To address this issue, Conditional GAN was proposed (Mirza & Osindero, 2014) using embedding with a huge penalty performance. Such an approach not only slowed down the training process but often made it difficult to fine-tune the parameters. Some other solutions were also proposed such as the use of Spectral Normalization (Miyato et al., 2018; Zhang et al., 2019) or Conditional Batch Normalization (de Vries et al., 2017; Dumoulin et al., 2016), with better performance in conditioning the noise vector. Ultimately, these additions provided the path for using labels to switch between classes. Unlike StyleGAN (Karras et al., 2019), where conditioning was addressed by using the Adaptive Instance Normalization (AdaIN) and a mapping network, this new method enabled intermediate results based on tiny variations to change the final output, resulting in a more flexible model. Despite the effectiveness of the new method, it was later modified in StyleGAN2 (Karras, Laine, et al., 2020) and StyleGAN2-ada (Karras, Aittala, et al., 2020), to address the blob-shaped that appeared in the images. For these two new architectures, the AdaIN processes separated the operations of normalization and modulation. Further details are offered in (Karras, Laine, et al., 2020).

While StyleGAN and StyleGAN2 were mainly focused on the quality of the results, only a limited number of the datasets were large enough to pro-vide good results. This was addressed by StyleGAN2-ada, which provides a data augmentation pipeline to avoid discriminator overfitting when working with small data (Karras, Aittala, et al., 2020), even scoring a lower FID. Others suggested image embedding (Abdal et al., 2019) to expand the latent space and augment the variety and number of the generated outcomes.

The latest direct modification in this path was introduced in StyleGAN3 (Karras et al., 2021) which introduced an alias-free approach to the architecture explored in Stylegan2-ada, the main goal behind these modifications was to detach the correlation between the generative process and the coordinates of the pixels to offer an optimized output for animations and videos while keeping the same FID score as its predecessor.

In addition to the improvements extended to the quality of the output, several streams modified the original algorithm for further applications, such as the extensible work introduced in the Generative Adversarial What-Where Network (Reed et al., 2016), where an approach was presented to generate conditioned images to locations. Another example is the case of SRGAN (Ledig et al., 2017) used to create super-resolution images. Further variations can also be seen in image-to-image translation, such as CycleGAN (Zhu et al., 2017), Pix2Pix (Isola et al., 2017a), and GANimorph (Gokaslan et al., 2018), these models aim to transfer features from one dataset to the other, of what is commonly known as style transfer. Currently, there exist other different approaches for image-to-image transformation, one that gained relevance recently has been Stable Diffusion models (Rombach et al., 2021) or Semantic Image Synthesis with Spatially-Adaptive Normalization (Park et al., 2019), but the core of these models is not based on the adversarial process, hence, out of the scope of this work.

Nevertheless, human-machine collaboration is deeply rooted in the domain and accelerated with the development of personal computing from the late seventies onward. A deeper review of the domain history reveals continuous demand for further human interactions within autonomous intelligent systems regardless of the apparent capabilities of machines (Janssen et al., 2019). The early work of (Carroll, 1997; Hoc, 2001; Laurance Rognin, 2000; Picard, 2003) discussed the affective and cognitive substance for this approach. Fundamentally, the development of generative models as such must remain human-centered in design, structure and application (Shneiderman, 2022). Hence, our work here continues the quest and aligns with a growing number of calls for similar collaborative approaches between humans and machines in light of recent advances in computational intelligence (Chignell et al., 2022; Dengel et al., 2021; German et al., 2019, 2020; Guo et al., 2020; Y. Yu et al., 2022; Zhuo, 2021). As character designers are placed in the loop of the proposed framework (See Figure 3), our main objective is to establish cognitive symbiotic (Inga et al., 2023) interactions between humans and machines (Rezwana & Maher, 2022), where machines' role is regarded as an instrument of knowledge magnification aiding in expanding perceptions, features, and connections that my exceeds human cognitive currency(Pasquinelli & Joler, 2020).

Therefore, an adaptation of different GAN architectures is used to develop a complementary framework that will cognitively aid character and conceptual designers at the initial stages of their projects by providing black/white silhouettes as visual propositions of concepts, followed by colored and textured concepts for the silhouettes representing different possible outcomes. The generated concepts here are intentionally visually suggestive, with different varieties and some randomness aiming to augment designers' agency during the creative process, by providing contextual visual affordance to motivates scaffolding (Estany & Martínez, 2014) and integration into new propositions.

## 4. Datasets and Models.

GANs create a latent space that can be freely explored after training to create new images different from the original data, allowing for flexibility and randomness in the output. To leverage both, randomness and realism, we propose two different models shown in Figures 1 and 2, while the first model is targeting a basic silhouette output, the second one is employed to generate colored and textured alternatives for the generated or given silhouette.

In the first stage of the process, a random noise vector is fed into a neural network trained on our dataset **Characters_silhouettes** to generate a black and white silhouette, which is then evaluated by character designers for concept development. In the second stage, we use a different network pretrained on another dataset, **Characters_colored**, that will be used to color the previously generated silhouettes to further assist human designers to develop their concept.

Constructed in tandem as a human-in-the-loop framework illustrated in Figure 3, the novelty and quality of the outputs are mediated between the generated stimuli, designers' competencies, contextual requirements, and the designed concepts. Due to copyright limitations, we cannot share the colored dataset used to train the models in Figure 2, but we publicly share a colored set of characters called **Characters_colored_gen**, containing 6, 000 images generated by this work. Several types of GANs were adapted to our context and trained on our dataset to analyze the quality of outputs. The implementation compares training the models using limited computational capabilities versus using transfer learning to determine the best approach possible. The subsections below introduce the details of the datasets and the models in question.

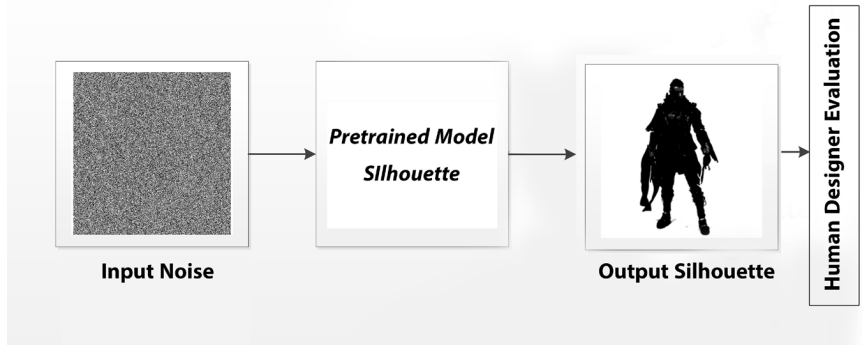

Figure 1: Randomly generated character by a noise vector.

*4.1. The Datasets.*

Most of the results explored in the GANs literature are limited to commonly used datasets, such as CIFAR10, MNIST, LSUN, or ImageNET. While they include similar human figures, these datasets are not focused on the features required for this context in terms of content, type, resolution, and number of images.

Our first dataset called **Characters_silhouettes** was used to train the GAN model deployed in the first stage (Figure 1), allowing the generation of silhouettes from random noise. The output of which is carried forward as an input for the second model deployed for the second stage (Figure 2), which is trained using a different dataset called **Characters_colored**, to generate colored versions for the black/white silhouettes.

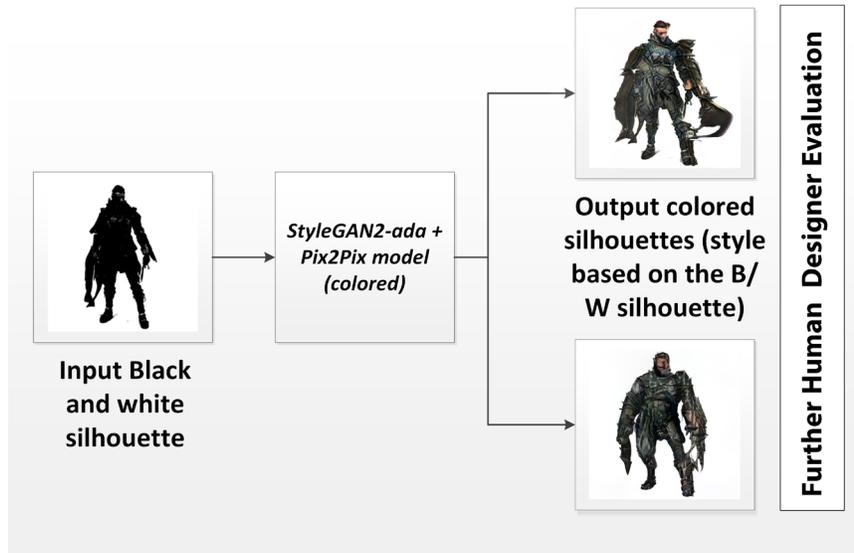

Figure 2: Colored generated character from a silhouette.

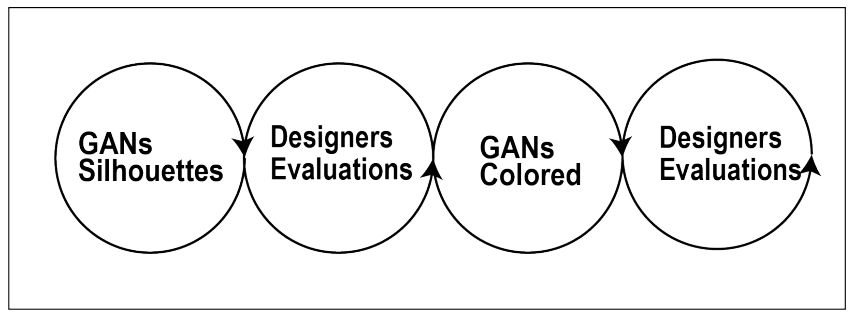

Figure 3: Co-creation process overview

**Characters_silhouettes:**

*Shape and resolution:* Squared images, resolution of 512×512. The original resolution of the images was no lower than 128×128 and they were upsampled using a bicubic filter.

*The number of images and labeling:* The set consists of 10k images split into 3 different classes called: Man, Monster, and Woman. As needed by some modules, the images were merged into a single class.

**Characters_colored:**

*Shape and resolution:* Squared images, resolution of 512×512. All images in this dataset were initially of a resolution of 512×512 or higher, they were downsampled when necessary.

*The number of images and labeling:* The set consists of 8.7k colored images and their respective silhouette version in black and white. Similar classes were used as per the first dataset.

*4.2. The Models.*

As we introduced at the beginning of this work, we trained two different consecutive models, one for the silhouettes, followed by another for the colored concepts. Generating silhouettes from random noise, as shown in Figure 1, is not as computationally demanding as generating colored images. Therefore, we evaluated the performance of the following models:

- Deep Convolutional GAN (DCGAN) (LeCun et al., 1998)

- Wasserstein GAN (WGAN) (Arjovsky et al., 2017)

- WGAN with Gradient Penalty (WGAN-GP) (Gulrajani et al., 2017)

- GAN-QP (Su, 2018)

- Large Scale GAN (BigGAN-deep) (Brock et al., 2019)

- Large Scale Adversarial Representation Learning (BiBIGGAN) (Donahue & Simonyan, 2019)

- StyleGAN2 with Adaptive Discriminator Augmentation (StyleGAN2-ada) (Karras, Aittala, et al., 2020)

For the second model (Figure 2), we combined the functionalities of Pix2Pix (Isola et al., 2017b) and StyleGAN2-ada (Karras, Aittala, et al., 2020) to color the silhouettes from the previous step, as well as add visual texture details to the final image. Both adapted models were trained using the **Characters_colored** dataset that was built using pairs of images with silhouettes and their respective colored versions. StyleGAN2-ada was trained by using the colored images only while Pix2Pix used the entire pair set. Both models were trained using the original image resolution of (512×512).

Because of the exhaustive details and fine-tuning iterations explored in this work, besides using conditional generators where possible to optimize the outcomes, another comparative study is underway that will explore different generative models including GANs, diffusion, and procedural generation. To remain focused on the scope discussed here, we refer the reader to the cited work of these models.

*4.3. Hardware and Software.*

GANs are powerful and computationally complex. Therefore, the state-of-the-art models are trained using a minimum of four powerful GPUs working in parallel. Even with such distribution, the training times are high, leading us to experiment using limited computational resources, at least for the first stage in our framework, to be able to demonstrate results and explore how different GAN architectures perform under these conditions. We also explore models' limitations in terms of dataset size and variation necessary for the desired output quality.

*4.3.1. Hardware.*

The first part of the experiment section is focused on showing the performance of GANs working on a computer with a single GPU. All the models, except for StyleGAN2-ada, were trained on Google Colab, a service that provides the equivalent of a computer with the following specifications (the exact specs could vary): CPU: Intel(R) Xeon(R) CPU @ 2.20GHz, GPU: K80 or T4 or P100, RAM: 12 GB.

The specs of the computer used for StyleGAN-ada and CycleGAN: CPU: Intel(R) Xeon(R) Gold 5120T CPU @ 2.20GHz, GPU: Quadro GV100 (32GB), RAM: 128 GB (only 32GB required for FID)

*4.3.2. Software.*

The software used depends on the model, StyleGAN2-ada, Pix2Pix and CycleGAN use Tensorflow while the others use Pytorch. The requirements for DCGAN, WGAN, WGAN-GP and BIGGAN-deep are: Pytorch 1.7.1, Torchvision 0.8.2, and CUDA 11.1.

The software packages used for StyleGAN2-ada, Pix2Pix and CycleGAN: Tensorflow 1.14, CUDA 10.0, cuDNN 7.5, Visual Studio 2015, and the VC Tools library are also required for StyleGAN2-ada.

The following section will dive into the details of the experiments performed in this work.

## 5. Experiments and Results.

As noted earlier, measuring the performance of GANs is not purely quantitative. A qualitative human expert evaluation is necessary to ensure that the results meet certain perceptual qualities to fulfill the aspired role. Furthermore, it is also crucial to consider the views of different experts in character design to better understand how variable design expertise interacts with the generated images toward creating new concepts. Hence, we structure the experiments section as follows:

- The quantitative section will present the detailed training process for different adapted models. If the results pass a short human review, a proper score is given for that model, with elaborations on the main features compared to other models as a clarification for a non-technical professional who may not work directly with Deep Learning models.

- Generating colored images from the silhouettes section presents the approach used to generate colored concepts to assist designers in visualizing possible outcomes.

- The qualitative section will introduce the work of participating designers who used the generated silhouettes for initial sketches, followed by a detailed view of the process used in such procedure to discuss the perceived advantages or otherwise difficulties faced while using the generated images as proposed in this work.

- An overview of the developed web application to interact with the best-performing models publicly, which will be later the subject of a longitudinal follow-up study.

*5.1. Quantitative Experiments.*

It is important to notice that most of the generators and discriminators have the same architecture, except for GAN-QP, BIGGAN BigBiGAN and StyleGAN2-ada, the layers of which are detailed in the models' section.

*5.1.1. Results for initial architectures (DCGAN, WGAN, WGAN-GP, GAN-QP)*

In this section we explore the results for some of the simplest GAN architectures such as DCGAN (Figure 4), WGAN (Figure 5), WGAN-GP (Figure 6) and GAN-QP (Figure 7), this will allow us to see the improvement when comparing them with more complex models.

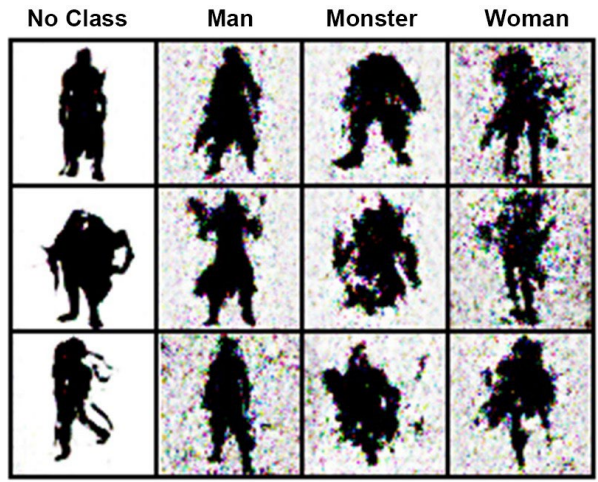

Figure 4: DCGAN generated images.

**General Observations:**

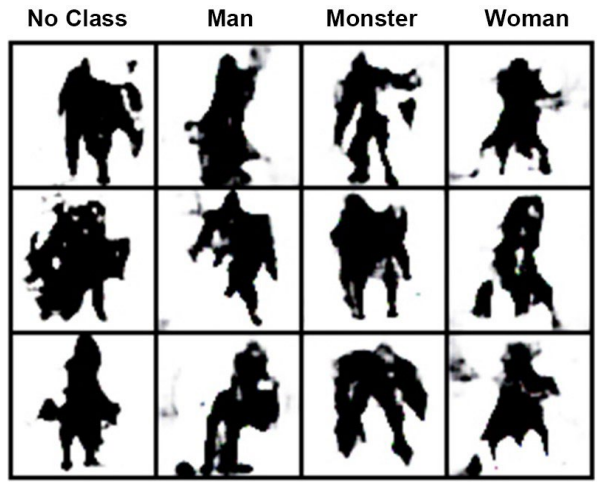

Figure 5: WGAN generated images.

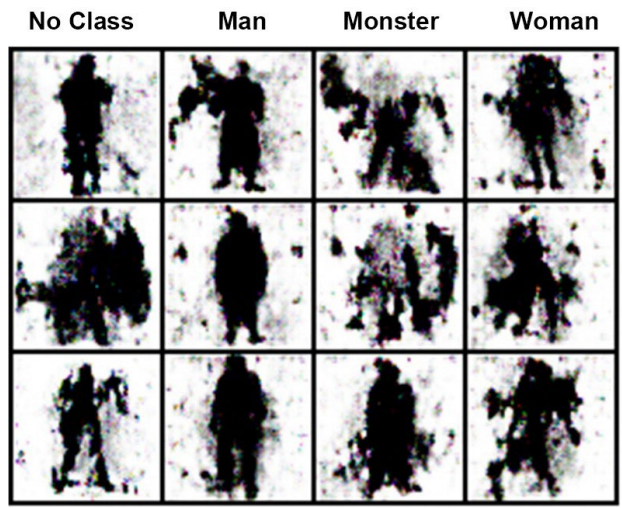

Figure 6: WGAN-GP generated images.

- Conditional models have a bad performance in general, this is due to how the images are conditioned. To overcome this issue, authors often adopt a similar strategy presented in Stylegan models.

- Wasserstein distance models perform much slower than regular models

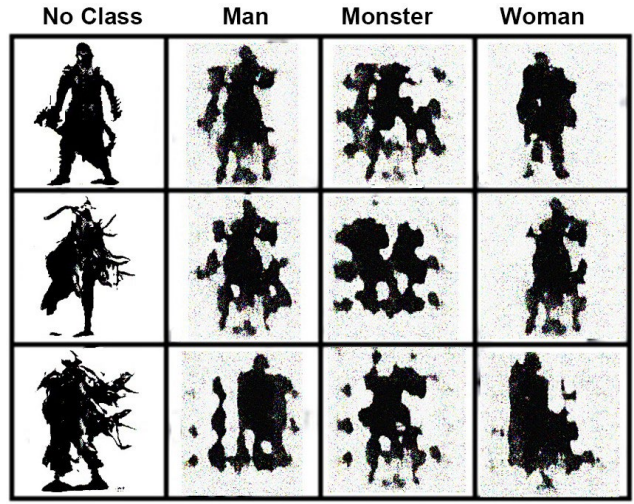

Figure 7: GAN-QP generated images.

due to the constraint they have, since they require to be 1-Lipchitz continuous, so the updates of the weights are also constrained.

– A big advantage for GAN-QP[2] is the possibility to easily handle images in higher resolutions, in our case, we directly trained the models using 512×512 images and the results are superior when compared to the other models.

*5.1.2. BIGGAN-deep and BigBiGAN*

The results for BIGGAN-deep and BigBiGAN are (only for the conditional version since it was designed with that purpose) shown in Figures 8 and 9, respectively, each row corresponds to a different class.

**Observations:**

– BIGGAN-deep represents an improvement in terms of quality but it is also the most challenging network to train on a single GPU since an appropriate batch size must be set initially as well as several parameters to be modified to obtain acceptable results.

---

[2]Pytorch implementation: https://github.com/RahulBhalley/gan-qp.pytorch

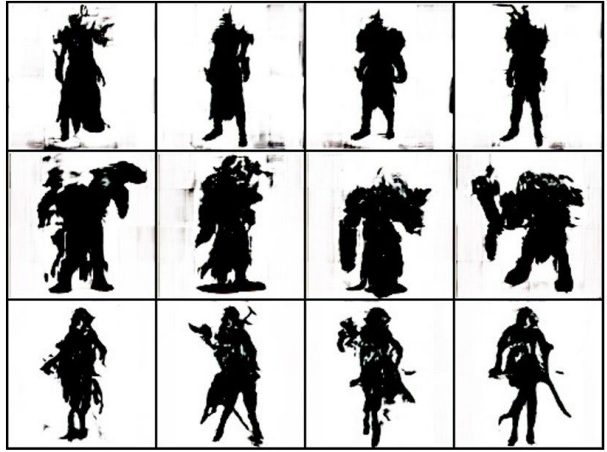

Figure 8: Conditional results for BIGGAN-deep.

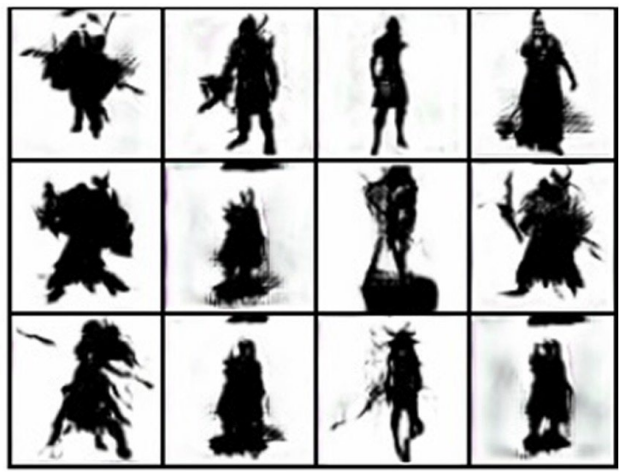

Figure 9: Conditional results for BigBiGAN.

- Mode collapse is significantly less frequent than observed in previous networks.

- BIGGAN-deep is the first network shown in this work, trained using a resolution of 128×128, replacing the 64×64 resolution formerly used by the previous models.

- The performance of BigBiGAN is similar to BIGGAN-deep, but it is

necessary to have more computational resources. In our case, we had to reduce the batch size and quality of the image to evaluate the model.

*5.1.3. StyleGAN2-ada (silhouettes)*

StyelGAN2-ada is one of the latest GAN architectures and its performance surpasses several of the models previously explored, but it also significantly increases training time and the GPU VRAM required.

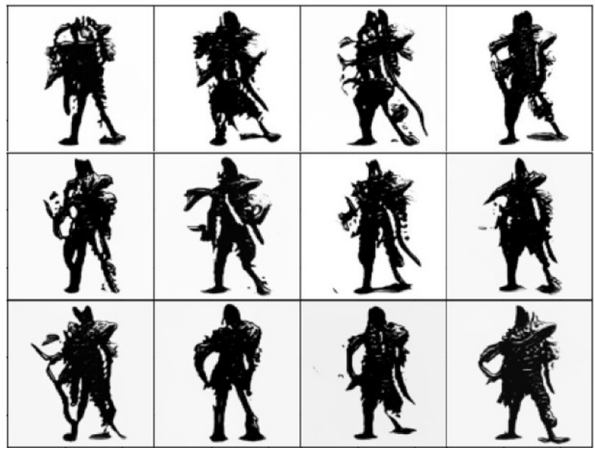

(a) Generated images obtained by a network trained from scratch (FID:105.69)

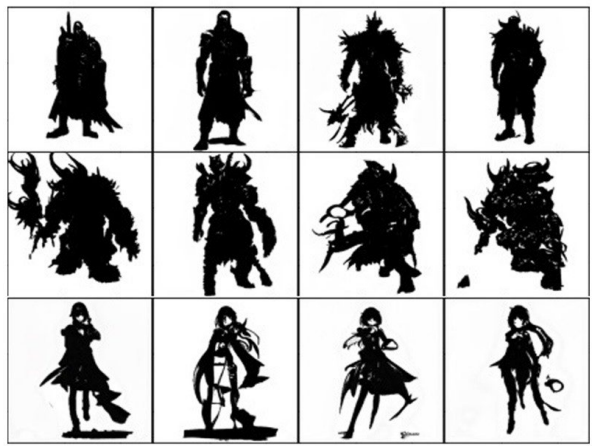

(b) Generated images obtained by a pretrained network (FID:17.60)

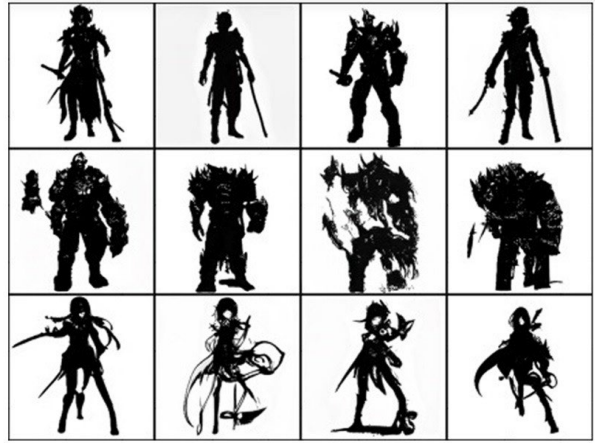

(c) Generated images obtained by a pretrained network (FID:17.53)

Figure 10: StyleGAN2-ada generated images.

Several runs were conducted with different criteria as follows:

a. Training without using any pretrained network (Figure 10a)

b. Training by using a pretrained network (ffhq1024, snap=10) (Figure 10b)

c. Training by using a pretrained network (ffhq1024, snap=10) and adding trunc=0.75 with the augmentation techniques (–augpipe=bgcfnc) (Figure 10c).

**Observations:**

– Training the model from scratch using our dataset still scored moderately compared to using transfer learning from the pretrained models. This is mostly due to the high computational resources and time required by the model to obtain such results.

– The results obtained using the pretrained networks are outstanding despite being trained on a different dataset (Tero Karras, n.d.) that doesn't share many similar features, however, this is reasonable since these snapshots of a model (shared usually as ".pkl" files or else known as pretrained pickles) were obtained after weeks using powerful GPUs running in parallel (8×TeslaV100).

*5.1.4. FID scores.*

In this subsection, we summarize all the FID scores obtained by the networks, which can be found in Table 1. To calculate the metric, we used 50K generated images by each network. For non-conditional networks, we generated 50K images without considering the class and calculated the FID for a merged version of our labelled subsets. As for the conditional GANs, we generated 50K images per class to calculate the metric. Nonetheless, to provide a comparison with their non-conditional counterparts, a total of 16.6K images per class were generated and merged to calculate the FID score. The code used to calculate the FID score is a Pytorch implementation (Seitzer, 2020).
As for the actual dataset generated for the FID, it can also be found as part of the publicly shared dataset at Mendeley Digital Commons Data (Lataifeh et al., 2022).

| GAN type | No class | Man | Monster | Woman |
| --- | --- | --- | --- | --- |
| DCGAN | 176.92 | N/A | N/A | N/A |
| WGAN | 71.25 | N/A | N/A | N/A |
| WGAN-GP | 112.59 | N/A | N/A | N/A |
| GAN-QP | 71.43 | N/A | N/A | N/A |
| condDCGAN | 224.87 | 199.36 | 254.58 | 264.66 |
| condWGAN | 135.93 | 127.97 | 160.82 | 149.57 |
| condWGAN-GP | 310.91 | 299.49 | 323.31 | 318.18 |
| condGAN-QP | 282.2 | 278.70 | 366.94 | 268.63 |
| BIGGAN-deep | 47.58 | 65.67 | 91.29 | 99.33 |
| BigBiGAN | 45.24 | 62.51 | 85.24 | 101.51 |
| StyleGAN2-ada (a) | 105.69 | N/A | N/A | N/A |
| StyleGAN2-ada (b) | 17.60 | N/A | N/A | N/A |
| StyleGAN2-ada (c) | 17.53 | N/A | N/A | N/A |

Table 1: FID scores for the models.

As listed in Table 1, some models are missing the scores for the class columns, since nonconditional models cannot generate images for each independent class. Additionally, the scores for CycleGAN and Pix2Pix were not calculated since a deeper comparison for image-to-image translation models is out of scope here.

*5.2. Generating Colored Images*

As we explored in the introduction section, a second output is provided consisting of colored images based on an original black-and-white silhouette. We propose a two-step solution to generate innovative approximations of the original silhouettes instead of just coloring them.

*5.2.1. Step 1: Coloring the Silhouettes*

To provide the best possible results, we explore two options: Pix2Pix and CycleGAN, both trained to transform a black-and-white silhouette into a colored one. The process is detailed next.

### *First Approach: Pix2Pix.*

Pix2Pix (Isola et al., 2017b) is an image-to-image model to transfer features from a set A of images into another set B with shared features. We propose to use this model to generate colored images for the silhouettes. The model at this stage was trained on another dataset called **Characters_colored** containing 8200 images of different classes with a resolution of 512×512 (not upsampled). We then tested the adapted model here on silhouettes generated by the StyleGAN2-ada model discussed earlier with the lowest FID score. The results can be seen in Figure 11, the images on the top row are the original silhouettes, compared to the output of the Pix2Pix model below.

The obtained results are of acceptable quality for the colored details, despite being produced based on a black-and-white silhouette. Nonetheless, further optimizations of colors, depth, and textures are still required to emphasize innovative aspects that are crucial for this work. Most of the generated colored shapes by Pix2Pix demonstrate little or no modification to the original draft. To enhance the quality and creativity of the colored generated images, we resort to an adaptation of StyleGAN.

### *Second Approach: CycleGAN.*

The aim for CycleGAN and Pix2Pix is similar. The only difference is the possibility to use the former bidirectionally, which allows for data transfer from a set A into another B and vice versa. The cycle iteration feature provides a significant value for different applications. However, as our main goal is to color the silhouettes and not to go back to the original black-and-white sample, only the coloring process will be considered in our analysis to determine which of the two approaches should be considered. Further detailed information on the process and architecture are provided in the original works (Isola et al., 2017a; Zhu et al., 2017).

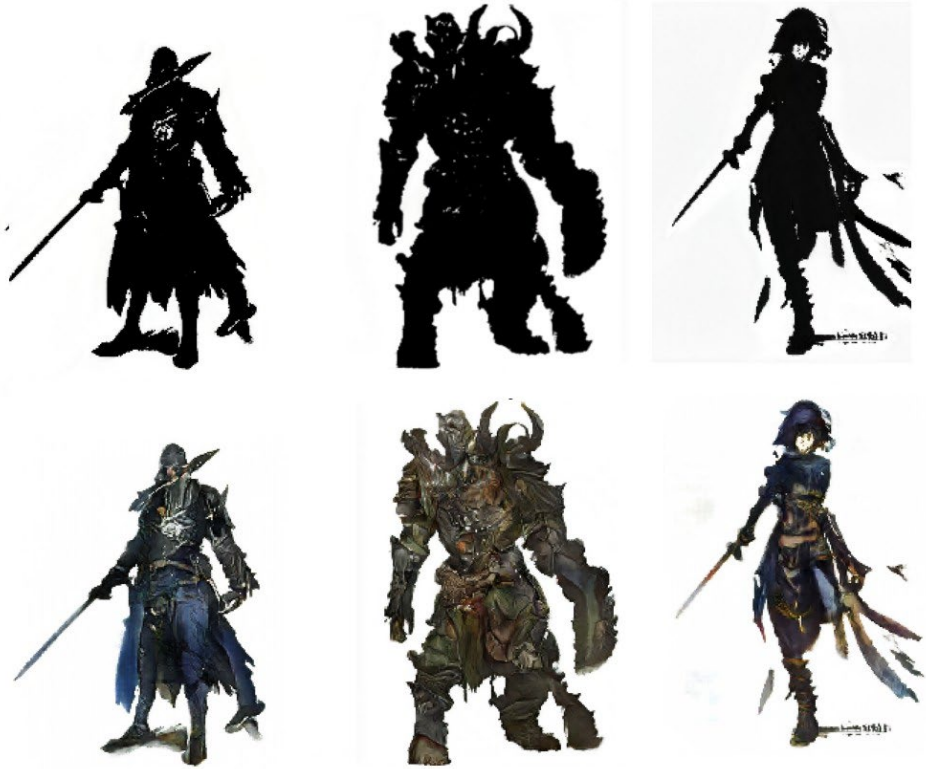

Figure 11: Colored output of the Pix2Pix model for generated images by StyleGAN2-ada.

The results seen in Figure 12 were generated by CycleGAN using the same original drafts that appeared in the Pix2Pix example in Figure 11. It is noticeable that the output images are not as good as the Pix2Pix output, the level of detail is lower with some images barely colored, which can be improved with further training, though, the same time was used for training both models.

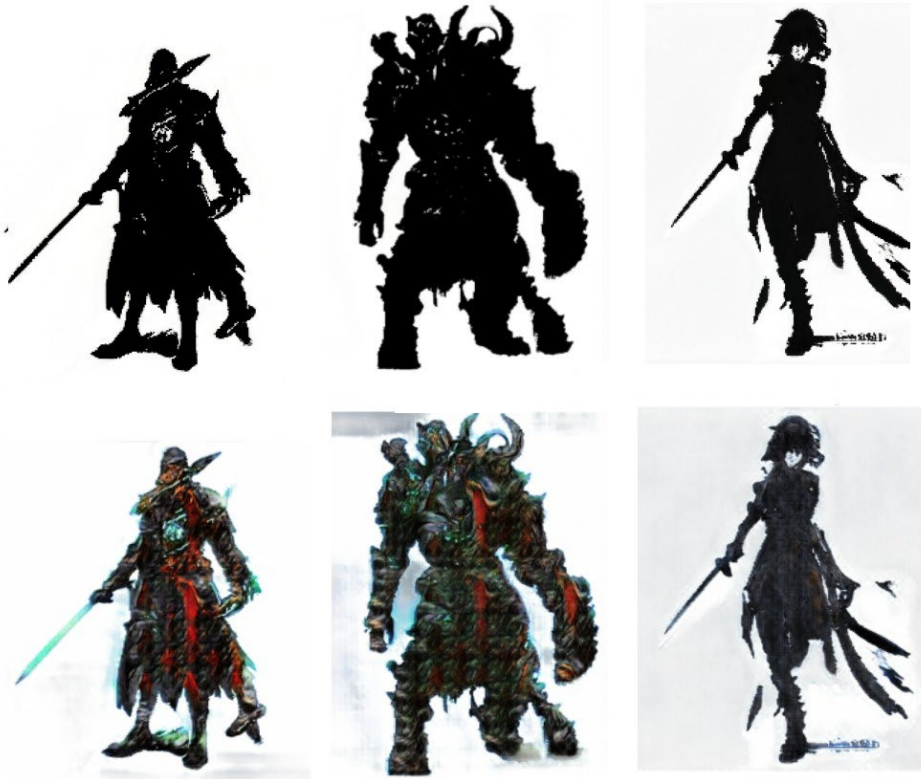

Figure 12: Colored output of the CycleGAN model for generated images by StyleGAN2-ada.

### 5.2.2. Step 2: StyleGAN2-ada

As part of the second approach, we also trained StyleGAN2-ada on the same dataset **Characters_colored** to project the colored diversity output of the models explored in the first step, which should allow us to obtain modified versions of these images without losing the main features. At this stage, we only used the generated images by Pix2Pix because of the quality and incoming information provided as input to the new model (StyleGAN2-ada), some samples of this process can be found in the image below, Figure 13.

The following section explores the externalized creative design process based on the visual enactment and vocal narrative of character designers as they use the presented framework to design new concepts.

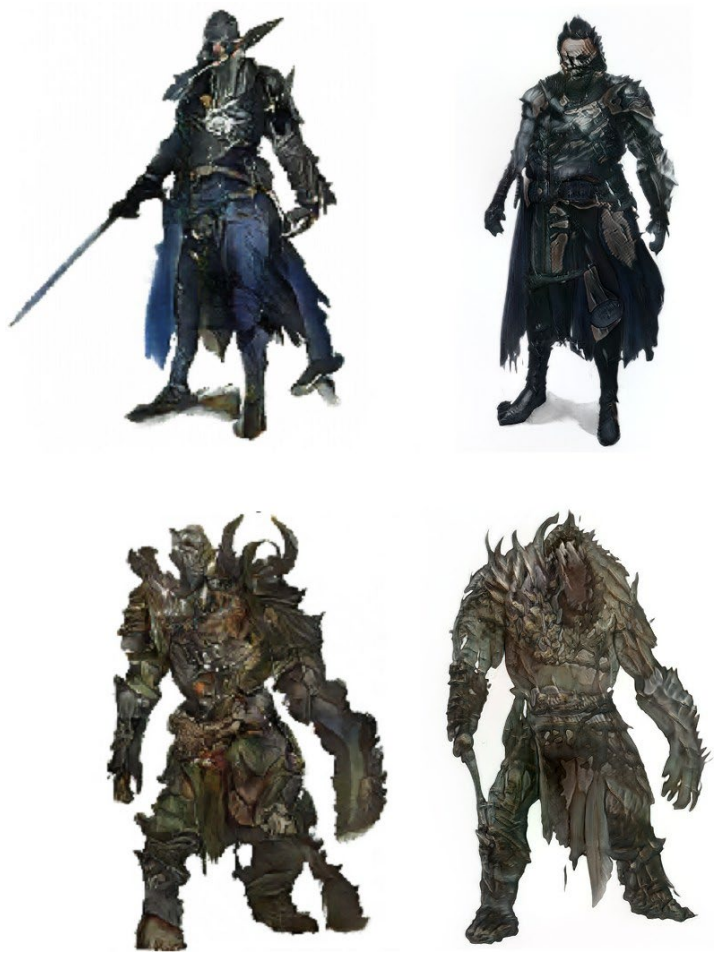

Figure 13: Results of StyleGAN2-ada trained on Characters_colored.

*5.3. Qualitative Exploration of the Creative Design Process*

In addition to the parametric evaluation, the generated outcomes were put to use within the proposed context to evaluate empirically their aspired value. Therefore, seven character designers were invited to participate in the evaluation process, five of whom agreed to collaborate remotely. The self-declared competencies came variable as novice, junior, intermediate, competent, and one expert designer recognized internationally in this domain. The characterization of these levels of expertise is further detailed in (Dreyfus & Dreyfus, 1986). The participant's designed concepts were fairly concordant with the reported competencies.

A pool of randomly generated silhouettes was shared with participants to use as a starting point for their concepts, but no further instructions or constraints were given concerning materials, style, or fidelity. The presented results below explore the variability of commitment, use, and adaptation of the silhouettes observed during the presented work. Post-work interviews were conducted with participants to clarify, and often, elaborate on some of the points highlighted from recorded notes using the think-aloud protocol (Payne, 1994).

Participating designers approached the visual silhouettes as that of a visual design brief, where an initial metaphoric, vague representation is drafted (Sadowska & Laffy, 2017). Hence, accelerating the design process observed and discussed by many (Dorst & Cross, 2001; Lawson, 1980; Nelson & Stolterman, 2003), moving from a blank canvas into a formulated representation with what is often described as a complex goal-oriented enactment (Visser, 2009). While further studies are needed to understand the influence of different cognitive abilities on observed behaviours (Vuong et al., 2022), the participants in this experiment found themselves in an ongoing conversation with externalized signs of a concept, a step closer to a concept that is yet to be conceived.

The different levels of expertise were certainly apparent in the fluency of such conversation, adding, moving, removing, framing, reflecting, and reframing (Schön, 1983) until the emergence of a new concept. Certainly, the design process continues to attract more interest in light of mediated computational capabilities, such as the one discussed in this work. Participating designers were often moving between the silhouette selected as a starting point and what they could imagine as possible. Visual designers do indeed see much more than what can be observed by others as mere visual sketches in space. This kind of seeing (Schön & Wiggins, 1992) is what set the motion forward for a dialogue with shapes, figures, directions, scales, and composition. Such a dialectical conversation (Goldschmidt, 1991) is evident in figures 14 to 18 below, with higher resolution samples added to appendix A. See Figures from A.21 to A.27.

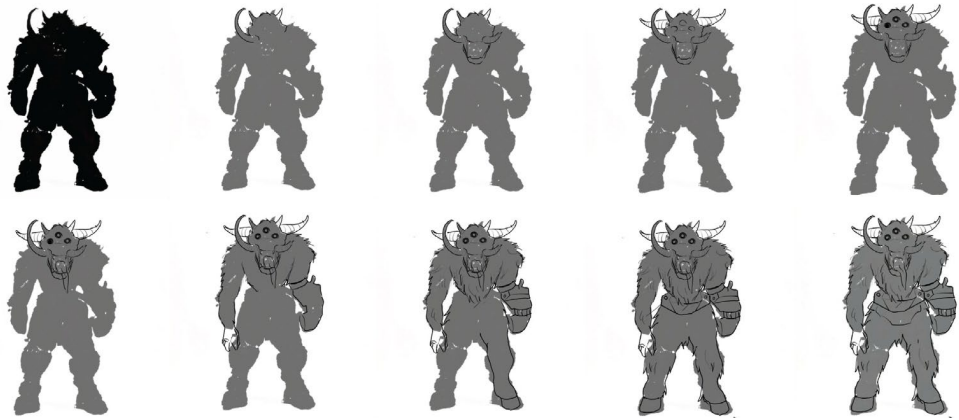

Figure 14: Concept development from silhouette - Novice.

We start with a sample taken from the novice character designer. While exhibiting fair artistic and anatomical competencies, the presented frames in Figure 14 demonstrate a faithful commitment to the selected silhouette, a behavior that falls within the characterization of beginners. Designers find it easier (at this level) to remain within the perceived rules, which may have been seen here as the silhouette outline. Consequently, the designer ventured within the silhouettes for clues, according to which, details were added, but rarely a re-formulation of the visual proposition. Which as a result, may risk becoming predictable, familiar, and possibly boring. The participant felt more confident that adding higher fidelity to the inner details will deliver the aspired or imagined result. During this conceptualization, the designer was motivated by the need to reinvent what she thought to exist within. The designer compared the process with that of sculpting within, suggesting that: *" I find myself pleased with the given figure, if I accept the proportions and the pose, I know I can find it within. Is that sculpting in 2d? Maybe. What I know for sure is that decoding visual clues from implicit to explicit scope"*. The concept, despite the self-imposed commitment to the boundaries, is unique and far different from machine-generated concepts.

Some aspects of the design process noted above persisted with the work of the intermediate designer as well. The illustrated concept progressions in Figures 15 and 16 demonstrate much more confidence in constructing beyond, across, and within the silhouette. The designer was adding, changing, and integrating different props. However, such concatenation came sequentially after the inward discovery and detailing of customs, features, and proportions. The designer's work here to an extent externalized the cognitive process of perceiving, drawing, pausing, re-framing, and reflecting. Although the description might appear simple, when considering the verbal narrative of the designer saying *"am not sure what is the plan here, if any, it is revealed to me with every stroke taken. I see something new, maybe it is difficult to explain, but I know am getting there."*. Undeniably, the mere definition of the emergence of visual concepts relates to different thoughts and ideas that may not be anticipated or planned before sketching, but morph through cycles of reinterpretations where images in the mind's influence sketched concepts (Menezes & Lawson, 2006).

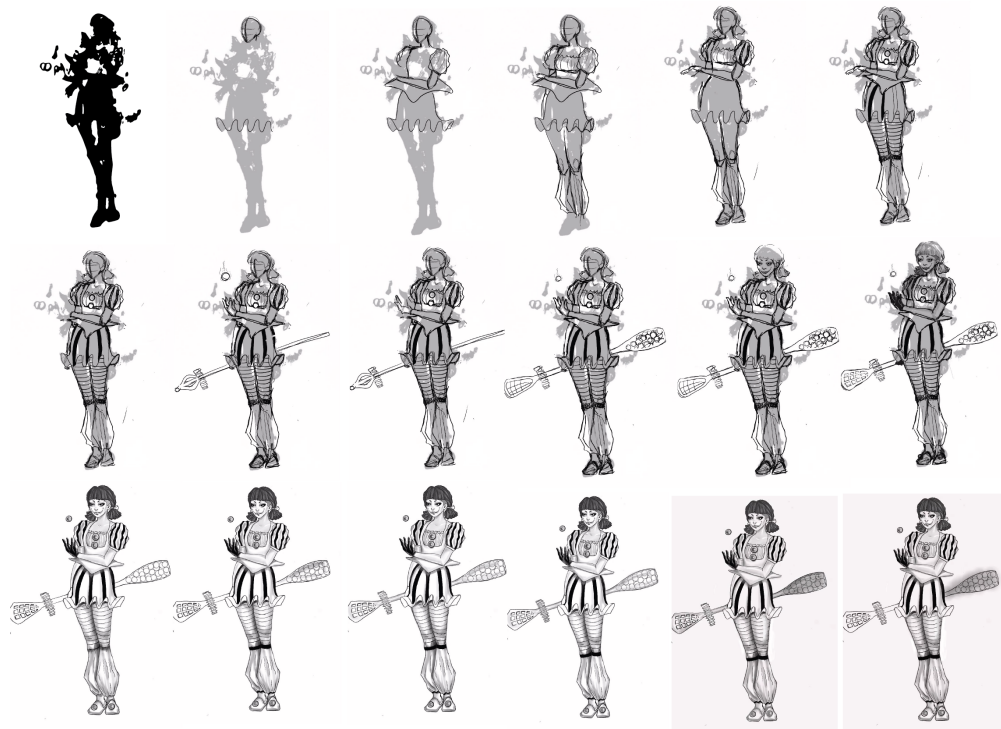

Figure 15: Concept development from silhouette - Intermediate

Unlike the previous participants, the discourse observed along with the explained cognitive process described by the expert designer as *transfiguring* from one vision into another. The frames depicted in Figures 17 and 18 are moving away from the initial visual proposition. The designer explains:" *My initial selection is based on visual curiosity. Then what am doing is simply a combination of intuition and visual clues revealed with every move. I do respect the proposed as a starting point, but I need to put it into a series of flipping, rotating, scaling, adding, removing, juxtaposing, and so on, until I see it there, ready to weave a story together*". Analyzing the progressions of these steps reveals an active visual dialogue that could not be settled without evoking the same figure boundaries that were perceived as restrictive written instructions for less competent designers. The emerging concept, as the designer maintained here, needs to be unmasked out of this vagueness, and for that, the visual journey is multi-directional inward, outward, within, and beyond the given visual boundaries of a silhouette. Such multi-directional process if we think about it, is indeed mimicked by several algorithms in machine learning models.

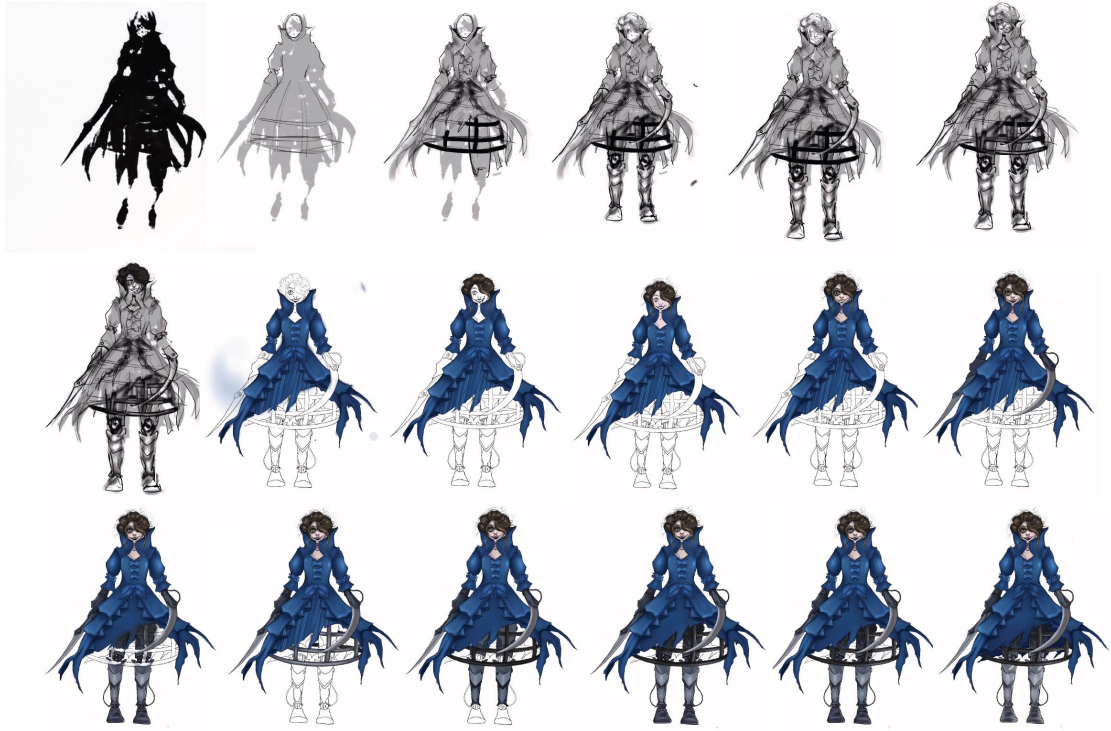

Figure 16: Concept development from silhouette - Intermediate

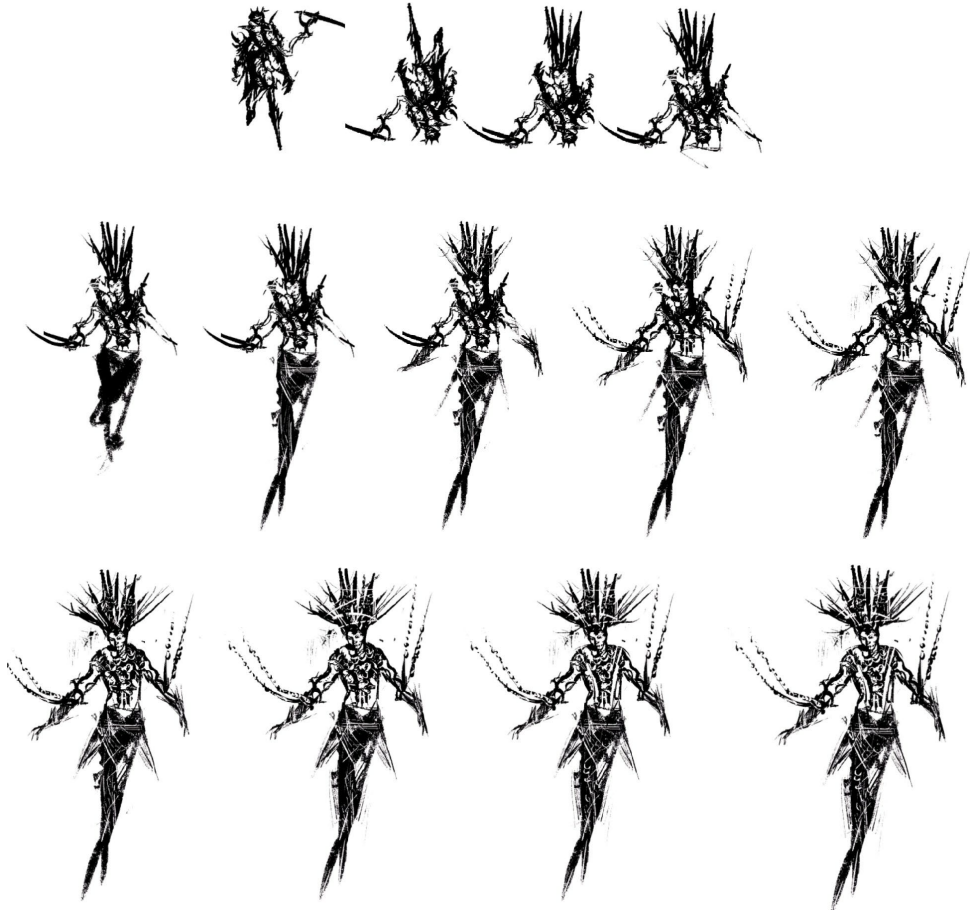

Figure 17: Concept development from silhouette - Expert.

*5.4. Web Application*

To engage with a wide community of concept designers across different domains, we finally propose a web utility to interact with our trained models, where users can create characters using either random generation or guided by an image as a base to follow. The application was built using the same process explained in this work. During the initial testing, we hosted the platform on a private cloud due to the model requirements, but it is being transferred to a public cloud. The web application will be expanded to explore other domains in the future such as landscapes, portraits, and buildings. The initial features available in the API are the following:

- *Random generation.* Despite obtaining a randomized output, we ensure the best possible result by providing a projection of a random image.

- *Guided generation.* Designers may upload an image and convert it into a new silhouette or colored concept with similar features.

- *Latent space exploration.* For both options offered in the web application, designers are encouraged to explore how their images can be modified by changing some parameters either randomly or using our the guided generation.

**6. Discussion.**

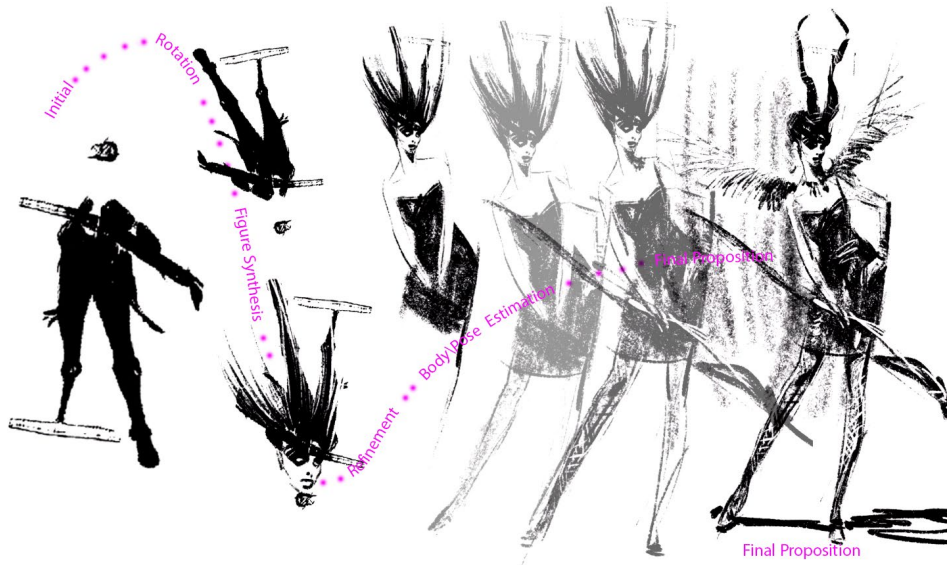

Figure 18: Concept development from silhouette - Expert.

This work was set to design and implement a novel framework for character design conceptualization integrating machine intelligence, GANs in particular, as a cognitive scaffolding that helps augment human creativity. The process of doing so, as detailed in the previous section, required several intermediate and cumulative visual dialectical actions performed by concept designers toward the creation of novel concepts.

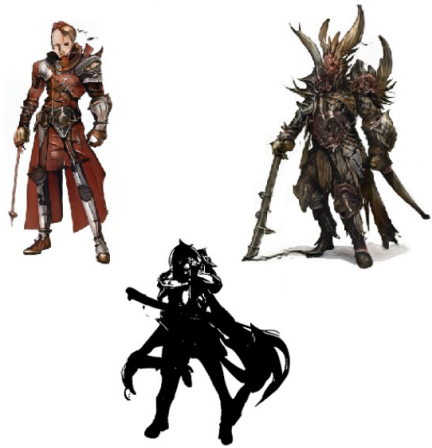

Figure 19: Random Character Generation Menu.

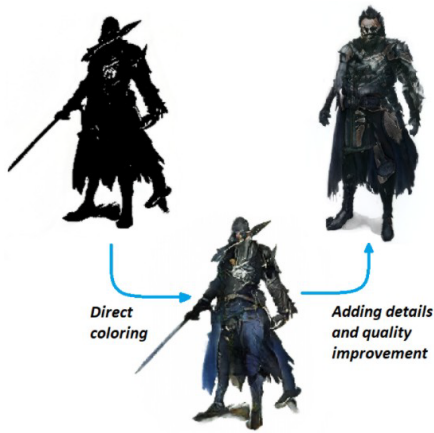

Figure 20: Silhouette to Character Generation Menu.

The second objective was also to demonstrate several strategies to overcome numerous constraints related to GANs' deployment with limited GPU resources. Therefore, we implemented and analyzed several GANs architectures to evaluate the appropriate one for such an application. Indeed, modern architectures provide incredible performance, albeit commanding costly resources that not all researchers in the domain can afford. Therefore, the networks used were fine-tuned with different adaptations to resolve the hardware drawback, while still meeting the quality and resolution parameters required for this work.

Unlike the mainstream application of GANs, the scope here does not aim to create hyper-realistic images, but rather a visual agent acting as a cognitive anchor to intrigue, direct, and inspire character designers to create novel concepts. Furthermore, a deeper analysis of the participating designer process externalized by their visual enactment and vocal narrative as they proceed affirms that machines-generated concepts, regardless of fidelity, were never seen more than cognitive substances in a visual dialogue led by humans. Such a view explains the reactions noted earlier toward generative models (Growcoot, 2023). Regardless of the quality of the generated work, the raised concerns are of serious ethical and moral implications (Dengel et al., 2021), and while we embrace computational intelligence, we see its role and place within a dialectical creative framework as proposed in this work.

The internal evaluation took into consideration two main parameters: output quality (resolution, variety, and novelty) and network training effectiveness. As a result, several networks were directly discarded for the lack of diversity or low-resolution images, such as the case of DCGAN and BIGGAN-deep, both of which excelled in other aspects mostly related to computational performance. Upon the presented exploration of several GANs, the selected model was StyleGAN2-ada. The outcome was first evaluated using the FID score calculated at this stage to evaluate performance on a new dataset. The obtained scores conveyed the effectiveness of the model, the adaptation of which proved useful for the specified creative process.

While GANs are critical to the success of the proposed collaborative approach, the scope was not to design a new model but to evaluate how and where in the design process can machine intelligence serve this creative domain. The adapted models integrated into the presented framework provided high levels of visual innovation compared to the initial training dataset. Despite being possible to train most of the models with limited resources, obtain-ing state-of-the-art results demand the use of several modern GPUs, capabilities that are not easily afforded by many researchers, but can alternatively be overcome using transfer learning techniques.

Nonetheless, we strongly encourage character designers, concept designers, and digital artists to consider GANs as part of their design process, for the cognitive scaffolding that can inspire and augment spatial and volumetric curiosity. We conclude that modern GAN architectures can perform well on the custom-built dataset for this particular context. Transfer learning is recommended to avoid hardware limitations.

Furthermore, the qualitative evaluation of the designers' outputs also affirmed the capabilities and utility of the proposed framework. The style, use, and adaptation of the generated silhouettes were concordant with established notions related to design expertise. Novice and intermediate designers were within the characteristics of their levels(Dreyfus & Dreyfus, 2005) as they remained close to the spatial boundaries of the proposed silhouette. Despite the self-imposed constraints, their output concepts were distinct from the provided silhouettes. However, expert designers who started with the generated concepts as a draft, quickly leaped into higher-order cognitive processes, demonstrated by leaps of imagination, confidence, and freedom from any implied restrictions of the design brief, symbolized here by the provided visuals, transforming a silhouette proposition to novel concepts.

Discussing the process with participating designers revealed an image of the internal cognitive process. For the novice and intermediate designers, the generated silhouette helped channel thoughts and direction of the quest, easing the initial stages of the process, crafting confidence in the materialized concept as being the answer to a puzzle, once found, designers felt exalt to expand on styles, textures, and tones to own the concept. Expert designers, however, saw much more than a visual clue while integrating part of the visual stimulus into a complex series of actions toward the creation of the final concept.

## 7. Conclusion.

The advancement of GANs witnessed over the last few years has extended their value and integration to a wide range of domains and purposes. This work has demonstrated the cognitive and creative value that can be provided by GANs to concept and character designers. The scope of this work is to propose a collaborative design process between humans and machines, where GANs generated concepts catalyze the design process that continues to be led by concept designers.

Several GANs were explored to evaluate their fit for such a creative process using both objectives (FID) and subjective measures (designers' review). Other than the outstanding FID scores, the results obtained have proven influential during the design process. While being approached as a visual design brief for beginners who demonstrated faithful commitment to shapes and figure boundaries, competent designers have seen much more in the generated outcome, the least of which, as an already ongoing creative visual dialogue that intrigued and incited deeper exploration for novelty. Furthermore, the designers' recorded process revealed a complex and highly creative process being externalized as designers moved from the initial perception of a visual value, followed by further actions that included adding, removing, pausing, rotating, re-framing, and reflecting for a novel concept to emerge.

We conclude that the proposed cognitive framework in this work has affirmed its value to the community, particularly as we move into a new era of immersive, extended, and mixed realities that continue to push the demands for new concepts, models, and most importantly, the methods to fulfill such demand. While the constructed dataset was limited to three classes to provide a proof of concept for the proposed, moving forward, this framework will be extended to include further elements, landscapes, objects, and animals. Additionally, the feedback that we expect from the larger community using the web-based utility noted earlier will certainly be of great value in disclosing needs, hopes, and future directions.

**Declaration of Competing Interest**

None.

**Acknowledgment**

The authors would like to acknowledge the valuable input of participating character designers, their interactions and integration of the model into their design process were instrumental to the validation of the work.

This research did not receive any specific grant from funding agencies in the public, commercial, or not-for-profit sectors.

**References**

Abdal, R., Qin, Y., & Wonka, P. (2019). Image2StyleGAN: How to embed images into the StyleGAN latent space? *Oct*, 4431–4440. https://doi.org/10.1109/ICCV.2019.00453

Arjovsky, M., Chintala, S., & Bottou, L. (2017). Wasserstein generative adversarial networks. *International conference on machine learning*, 214–223.

Baio, A. (2022). *Invasive diffusion: How one unwilling illustrator found herself turned into an ai model*. Waxy.org.

Barnard, P. J., & May, J. (1999). Representing cognitive activity in complex tasks. *Human–Computer Interaction*, *14*(1-2), 93–158. https://doi.org/10.1080/07370024.1999.9667267

Borji, A. (2019). Pros and cons of gan evaluation measures. *179*, 41–65. https://doi.org/10.1016/j.cviu.2018.10.009

Brock, A., Donahue, J., & Simonyan, K. (2019). Large scale GAN training for high fidelity natural image synthesis. *International Conference on Learning Representations*.

Carroll, J. M. (1997). Human-computer interaction: Psychology as a science of design. *Annual Review of Psychology*, *48*(1), 61–83. https://doi.org/10.1146/annurev.psych.48.1.61

Chignell, M., Wang, L., Zare, A., & Li, J. (2022). The evolution of hci and human factors: Integrating human and artificial intelligence. *ACM Trans. Comput.-Hum. Interact.* https://doi.org/10.1145/3557891

Creswell, A., White, T., Dumoulin, V., Arulkumaran, K., Sengupta, B., & Bharath, A. A. (2018). Generative Adversarial Networks: An Overview. *35*(1), 53–65. https://doi.org/10.1109/MSP.2017.2765202

Csikszentmihalyi, M. (1988). Society, culture, and person: a systems view of creativity. In R. Sternberg (Ed.), *The nature of creativity: Contemporary psychological perspectives* (pp. 325–339). Cambridge University Press.

Deng, J., Dong, W., Socher, R., Li, L.-J., Li, K., & Fei-Fei, L. (2009). Imagenet: A large-scale hierarchical image database. *2009 IEEE conference on computer vision and pattern recognition*, 248–255.

Dengel, A., Devillers, L., & Schaal, L. M. (2021). Augmented human and human-machine co-evolution: Efficiency and ethics. *Lecture Notes in Computer Science (including subseries Lecture Notes in Artificial Intelligence and Lecture Notes in Bioinformatics)*, *12600 LNCS*, 203–227. https://doi.org/10.1007/978-3-030-69128-8_13

de Vries, H., Strub, F., Mary, J., Larochelle, H., Pietquin, O., & Courville, A. (2017). Modulating early visual processing by language. *Proceedings of the 31st International Conference on Neural Information Processing Systems*, 6597–6607.

Donahue, J., & Simonyan, K. (2019). Large scale adversarial representation learning. In *Proceedings of the 33rd international conference on neural information processing systems*. Curran Associates Inc.

Dorst, K., & Cross, N. (2001). Creativity in the design process: Co-evolution of problem–solution. *Design studies*, *22*(5), 425–437.

Dorst, K., & Reymen, I. (2004). Levels of expertise in design education. *DS 33: Proceedings of E&PDE 2004, the 7th International Conference on Engineering and Product Design Education, Delft, the Netherlands*.

Dreyfus, H. L., & Dreyfus, S. E. (1986). From socrates to expert systems: The limits of calculative rationality. In C. Mitcham & A. Huning (Eds.), *Philosophy and technology ii: Information technology and computers in theory and practice* (pp. 111–130). Springer Netherlands. https://doi.org/10.1007/978-94-009-4512-8_9

Dreyfus, H. L., & Dreyfus, S. E. (2005). Peripheral Vision: Expertise in Real World Contexts. *Organization Studies*, *26*(5), 779–792. https://doi.org/10.1177/0170840605053102

Dumoulin, V., Shlens, J., & Kudlur, M. (2016). A learned representation for artistic style. *abs/1610.07629*.

Estany, A., & Martínez, S. (2014). "scaffolding" and "affordance" as integrative concepts in the cognitive sciences. *Philosophical Psychology*, *27* (1), 98–111. https://doi.org/10.1080/09515089.2013.828569

Fish, J., & Scrivener, S. (1990). Amplifying the mind's eye: Sketching and visual cognition amplifying the mind's eye: Sketching and visual cognition. *Leonardo*, *23*, 117–126.

Funke, C. M., Borowski, J., Stosio, K., Brendel, W., Wallis, T. S., & Bethge, M. (2019). The notorious difficulty of comparing human and machine perception. *Proceeding of the Conference on Cognitive Computational Neuroscience*, 642–646.

German, K., Limm, M., Wölfel, M., & Helmerdig, S. (2019). Towards artificial intelligence serving as an inspiring co-creation partner. *EAI Endorsed Transactions on Creative Technologies*, *6*(19).

German, K., Limm, M., Wölfel, M., & Helmerdig, S. (2020). Co-designing object shapes with artificial intelligence. *Lecture Notes of the Institute for Computer Sciences, Social-Informatics and Telecommunications Engineering, LNICST, 328 LNICST*, 309–327. https://doi. org/10 . 1007/978-3-030-53294-9_21

Gokaslan, A., Ramanujan, V., Ritchie, D., Kim, K. I., & Tompkin, J. (2018). Improving shape deformation in unsupervised image-to-image translation. *11216 LNCS*, 662–678. https://doi.org/10.1007/978- 3- 030- 01258-8_40

Goldschmidt, G. (1991). The dialectics of sketching. *Creativity Research Journal*, *4*, 123–143. https://doi.org/10.1080/10400419109534381

Goodfellow, I., Pouget-Abadie, J., Mirza, M., Xu, B., Warde-Farley, D., Ozair, S., Courville, A., & Bengio, Y. (2014). Generative adversarial nets. *27*, 2672–2680.

Growcoot, M. (2023). *Lawsuit filed against ai image generators stable diffusion and midjourney*. PetaPixel.

Gulrajani, I., Ahmed, F., Arjovsky, M., Dumoulin, V., & Courville, A. C. (2017). Improved training of wasserstein gans. In I. Guyon, U. V. Luxburg, S. Bengio, H. Wallach, R. Fergus, S. Vishwanathan, & R. Garnett (Eds.), *Advances in neural information processing systems*. Curran Associates, Inc.

Guo, C., Bai, T., Lu, Y., Lin, Y., Xiong, G., Wang, X., & Wang, F.-Y. Skywork-davinci: A novel cpss-based painting support system. In: *2020-August*. 2020, 673–678. https://doi.org/10.1109/CASE48305.2020. 9216814.

Heusel, M., Ramsauer, H., Unterthiner, T., Nessler, B., & Hochreiter, S. (2017). Gans trained by a two time-scale update rule converge to a local nash equilibrium, 6629–6640.

Hinton, G., Vinyals, O., & Dean, J. (2015). Distilling the knowledge in a neural network. https://doi.org/10.48550/ARXIV.1503.02531

Hoc, J.-M. (2001). Towards a cognitive approach to human–machine cooperation in dynamic situations. *International Journal of Human-Computer Studies*, *54*(4), 509–540. https://doi.org/https://doi.org/10.1006/ijhc.2000.0454

Hong, Y., Hwang, U., Yoo, J., & Yoon, S. (2019). How generative adversarial networks and their variants work: An overview. *52*(1), 1–43. https://doi.org/10.1145/3301282

Inga, J., Ruess, M., Robens, J. H., Nelius, T., Rothfuß, S., Kille, S., Dahlinger, P., Lindenmann, A., Thomaschke, R., Neumann, G., Matthiesen, S., Hohmann, S., & Kiesel, A. (2023). Human-machine symbiosis: A multivariate perspective for physically coupled human-machine systems. *International Journal of Human-Computer Studies*, *170*, 102926. https://doi.org/https://doi.org/10.1016/j.ijhcs.2022.102926

Isola, P., Zhu, J.-Y., Zhou, T., & Efros, A. A. (2017a). Image-to-image translation with conditional adversarial networks. *2017 IEEE Conference on Computer Vision and Pattern Recognition (CVPR)*, 5967–5976. https://doi.org/10.1109/CVPR.2017.632

Isola, P., Zhu, J.-Y., Zhou, T., & Efros, A. A. (2017b). Image-to-image translation with conditional adversarial networks. *Proceedings of the IEEE conference on computer vision and pattern recognition*, 1125–1134.

Janssen, C. P., Donker, S. F., Brumby, D. P., & Kun, A. L. (2019). History and future of human-automation interaction [50 years of the International Journal of Human-Computer Studies. Reflections on the past, present and future of human-centred technologies]. *International Journal of Human-Computer Studies*, *131*, 99–107. https://doi.org/https://doi.org/10.1016/j.ijhcs.2019.05.006

Jansson, D. G., & Smith, S. M. (1991). Design fixation. *12*(1), 3–11. https://doi.org/10.1016/0142-694X(91)90003-F

Karras, T., Aila, T., Laine, S., & Lehtinen, J. (2017). Progressive growing of gans for improved quality, stability, and variation. *abs/1710.10196*.

Karras, T., Aittala, M., Hellsten, J., Laine, S., Lehtinen, J., & Aila, T. (2020). Training generative adversarial networks with limited data. In H. Larochelle, M. Ranzato, R. Hadsell, M. F. Balcan, & H. Lin (Eds.), *Advances in neural information processing systems* (pp. 12104–12114). Curran Associates, Inc.

Karras, T., Aittala, M., Laine, S., Härkönen, E., Hellsten, J., Lehtinen, J., & Aila, T. (2021). Alias-free generative adversarial networks. *Proc. NeurIPS*.

Karras, T., Laine, S., & Aila, T. (2019). A style-based generator architecture for generative adversarial networks. *2019 IEEE/CVF Conference on Computer Vision and Pattern Recognition (CVPR)*, 4396–4405. https://doi.org/10.1109/CVPR.2019.00453

Karras, T., Laine, S., Aittala, M., Hellsten, J., Lehtinen, J., & Aila, T. (2020). Analyzing and improving the image quality of stylegan. *IEEE/CVF Conference on Computer Vision and Pattern Recognition (CVPR)*, 8107–8116. https://doi.org/10.1109/CVPR42600.2020.00813

Krizhevsky, A., Nair, V., & Hinton, G. (2006). Cifar-10 (canadian institute for advanced research). http://www.cs.toronto.edu/~kriz/cifar.html

Lataifeh, M., Carrasco, X., & Elnagar, A. (2022). Diversified character dataset for creative applications (dcdca). https://doi.org/10.17632/sdwbf4xrwz.1

Laurance Rognin, M. Z., Pascal Salemier. (2000). Cooperation, reliability of socio-technical systems and allocation of function. *International Journal of Human-Computer Studies*, *52*(2), 357–379. https://doi.org/https://doi.org/10.1006/ijhc.1999.0293

Lawson, B. (1980). *How designers think*. Architectural Press.

LeCun, Y., Bottou, L., Bengio, Y., & Haffner, P. (1998). Gradient-based learning applied to document recognition. https://doi.org/10.1109/5.726791

Ledig, C., Theis, L., Huszár, F., Caballero, J., Cunningham, A., Acosta, A., Aitken, A., Tejani, A., Totz, J., Wang, Z., & Shi, W. (2017). Photo-realistic single image super-resolution using a generative adversarial network. *Jan*, 105–114. https://doi.org/10.1109/CVPR.2017.19

Liao, J., Hansen, P., & Chai, C. (2020). A framework of artificial intelligence augmented design support. *Human–Computer Interaction*, *35*(5-6), 511–544. https://doi.org/10.1080/07370024.2020.1733576

Mahdizadehaghdam, S., Panahi, A., & Krim, H. (2019). Sparse generative adversarial network. *Proceedings - 2019 International Conference on Computer Vision Workshop, ICCVW 2019*, 3063–3071. https://doi.org/10.1109/ICCVW.2019.00369

McFarlane, D. C., & Latorella, K. A. (2002). The scope and importance of human interruption in human-computer interaction design. *Human–*

*Computer Interaction*, *17* (1), 1–61. https://doi.org/10.1207/S15327051HCI1701\_1

Menezes, A., & Lawson, B. (2006). How designers perceive sketches. *Design Studies*, *27*, 571–585. https://doi.org/10.1016/j.destud.2006.02.001

Mirza, M., & Osindero, S. (2014). Conditional generative adversarial nets. *abs/1411.1784*.

Miyato, T., Kataoka, T., Koyama, M., & Yoshida, Y. (2018). Spectral normalization for generative adversarial networks. *International Conference on Learning Representations*.

Molich, R., & Nielsen, J. (1990). Improving a human-computer dialogue. *33*(3), 338–348. https://doi.org/10.1145/77481.77486

Nelson, H., & Stolterman, E. (2003). *The design way* (1st). Educational Technology Publications.

Park, T., Liu, M.-Y., Wang, T.-C., & Zhu, J.-Y. (2019). Semantic image synthesis with spatially-adaptive normalization. *2019 IEEE/CVF Conference on Computer Vision and Pattern Recognition (CVPR)*. https://doi.org/10.1109/cvpr.2019.00244

Pasquinelli, M., & Joler, V. (2020). The nooscope manifested. *AI as Instrument of Knowledge Extractivism (https://nooscope. ai/)*.

Payne, J. W. (1994). Thinking aloud: Insights into information processing. *5*(5), 241–248. https://doi.org/10.1111/j.1467-9280.1994.tb00620.x

Picard, R. W. (2003). Affective computing: Challenges. *International Journal of Human-Computer Studies*, *59*(1-2), 55–64.

Reed, S., Akata, Z., Mohan, S., Tenka, S., Schiele, B., & Lee, H. (2016). Learning what and where to draw. *Proceedings of the 30th International Conference on Neural Information Processing Systems*, 217–225.

Rezwana, J., & Maher, M. L. (2022). Designing creative ai partners with cofi: A framework for modeling interaction in human-ai co-creative systems. *ACM Trans. Comput.-Hum. Interact.* https://doi.org/10.1145/3519026

Rombach, R., Blattmann, A., Lorenz, D., Esser, P., & Ommer, B. (2021). High-resolution image synthesis with latent diffusion models. https://doi.org/10.48550/ARXIV.2112.10752

Sadowska, N., & Laffy, D. (2017). The design brief: Inquiry into the starting point in a learning journey. *Design Journal*, *20*, S1380–S1389. https://doi.org/10.1080/14606925.2017.1352664

Schön, D. A. (1983). *The reflective practitioner: How professionals think in action* (1st). Basic Books.

Schön, D. A., & Wiggins, G. (1992). *Kinds of seeing and their functions in designing*.

Seitzer, M. (2020). pytorch-fid: FID Score for PyTorch [Version 0.3.0]. %5Curl%7Bhttps://github.com/mseitzer/pytorch-fid%7D

Shneiderman, B. (2022). *Human-centered ai*. Oxford University Press.

Su, J. (2018). Gan-qp: A novel gan framework without gradient vanishing and lipschitz constraint. *ArXiv*, *abs/1811.07296*.

Tero Karras, T. A., Samuli Laine. (n.d.). Flickr-Faces-HQ Dataset (FFHQ), year=2018, url=https://github.com/nvlabs/ffhq-dataset,

Theis, L., van den Oord, A., & Bethge, M. (2015). A note on the evaluation of generative models.

Visser, W. (2009). Design: One, but in different forms. *Design studies*, *30*(3), 187–223.

Viswanathan, V., & Linsey, J. (2012). A study on the role of expertise in design fixation and its mitigation. *Proceedings of the ASME Design Engineering Technical Conference*, *7*. https://doi.org/10.1115/DETC2012-71155

Vuong, T., Jacucci, G., & Ruotsalo, T. (2022). Naturalistic digital behavior predicts cognitive abilities. *The ACM CHI Conference on Human Factors in Computing Systems*.

Weatherbed, J. (2022). *Artstation is hiding images protesting ai art on the platform*.

WHO. (2019). Burn-out an "occupational phenomenon": International classification of diseases. Retrieved February 19, 2021, from https://www.who.int/news/item/28-05-2019-burn-out-an-occupational-phenomenon-international-classification-of-diseases

Yu, F., Zhang, Y., Song, S., Seff, A., & Xiao, J. (2015). LSUN: construction of a large-scale image dataset using deep learning with humans in the loop. *CoRR*, *abs/1506.03365*. http://arxiv.org/abs/1506.03365

Yu, Y., Yu, H., Cho, J., Park, J., Lim, E., & Ha, J. Human-ai co-creation practice to reconfigure the cultural emotion : Han. In: 2022, 414–417. https://doi.org/10.1145/3524458.3547127.

Zhang, H., Goodfellow, I., Metaxas, D., & Odena, A. (2019). Self-attention generative adversarial networks. In K. Chaudhuri & R. Salakhutdinov (Eds.), *Proceedings of the 36th international conference on machine learning* (pp. 7354–7363). PMLR.

Zhu, J.-Y., Park, T., Isola, P., & Efros, A. A. (2017). Unpaired image-to-image translation using cycle-consistent adversarial networks. *2017 IEEE International Conference on Computer Vision (ICCV)*, 2242–2251. https://doi.org/10.1109/ICCV.2017.244

Zhuo, F. Human-machine co-creation on artistic paintings. In: 2021, 316–319. https://doi.org/10.1109/DTPI52967.2021.9540122.

# Appendix A. Images in high Resolution

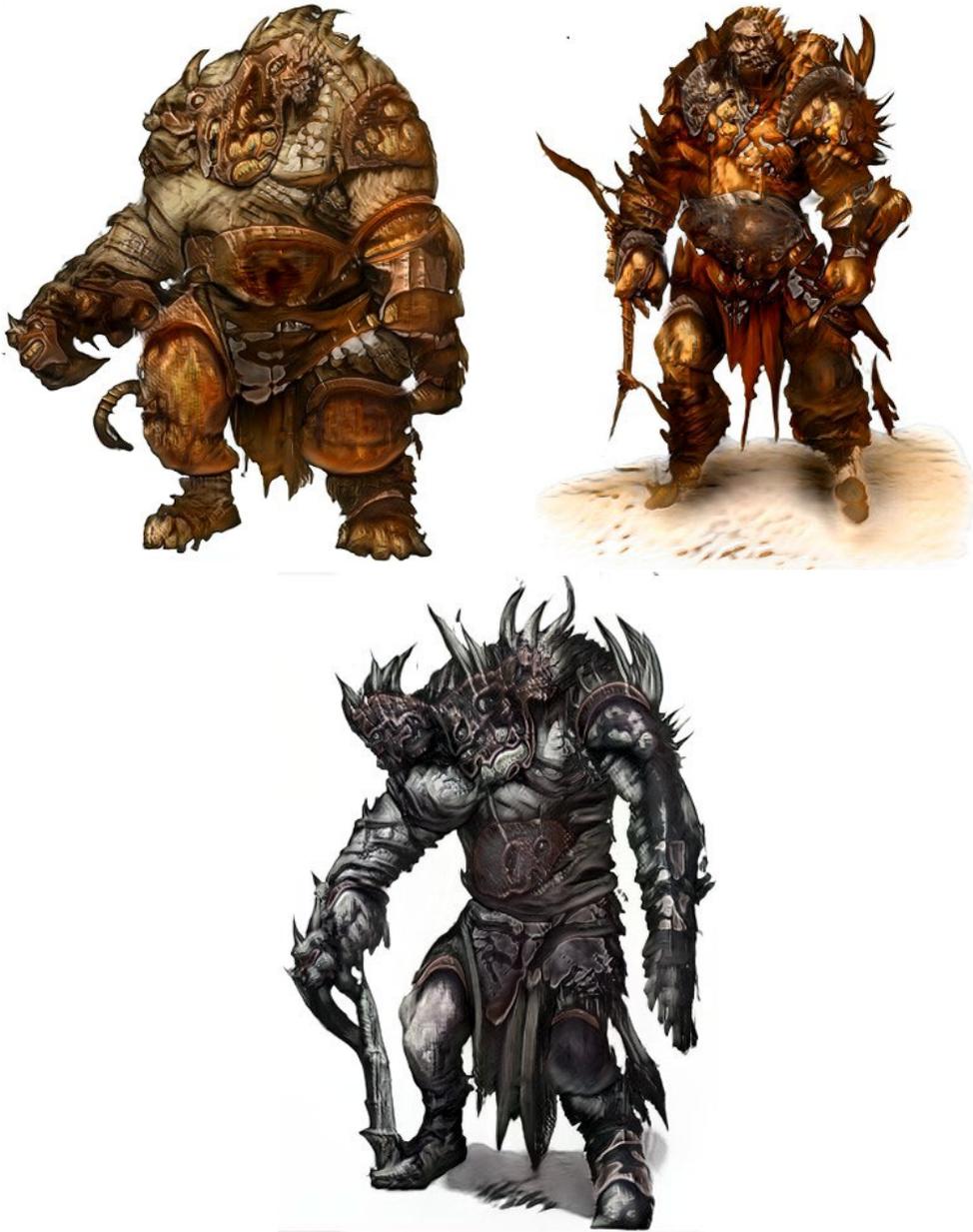

Figure A.21: Monsters - Generated images

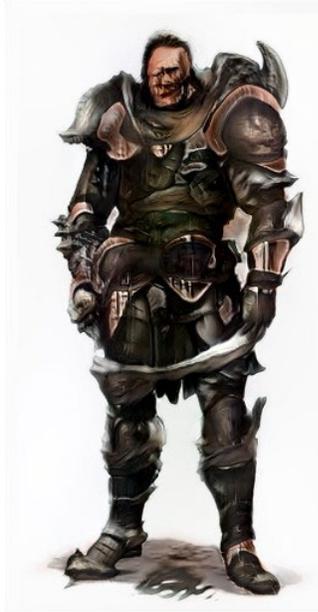
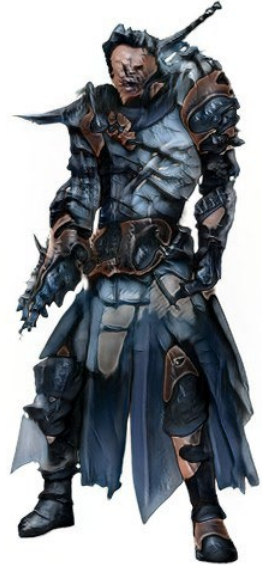
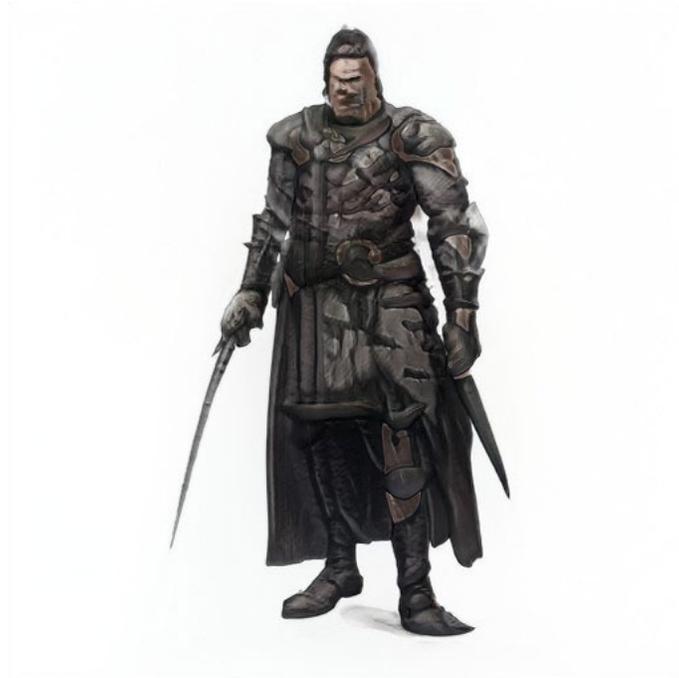

Figure A.22: Men - Generated images

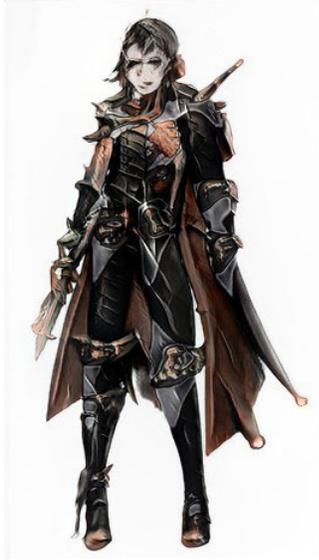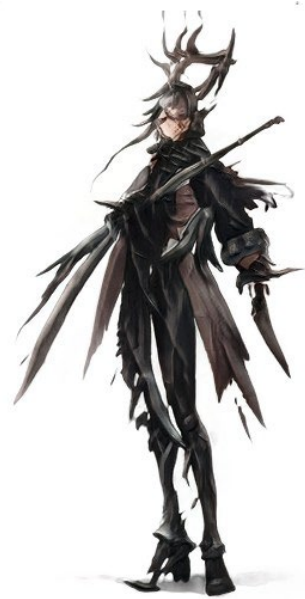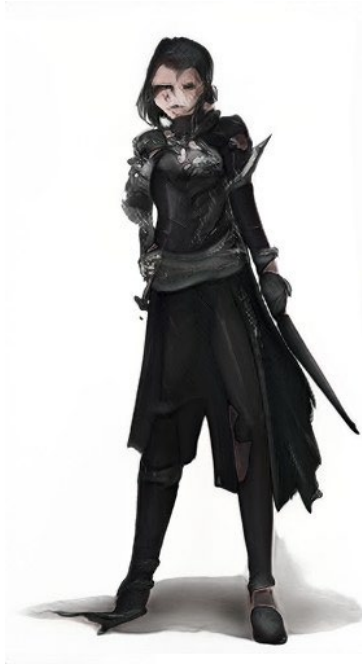

Figure A.23: Women - Generated images

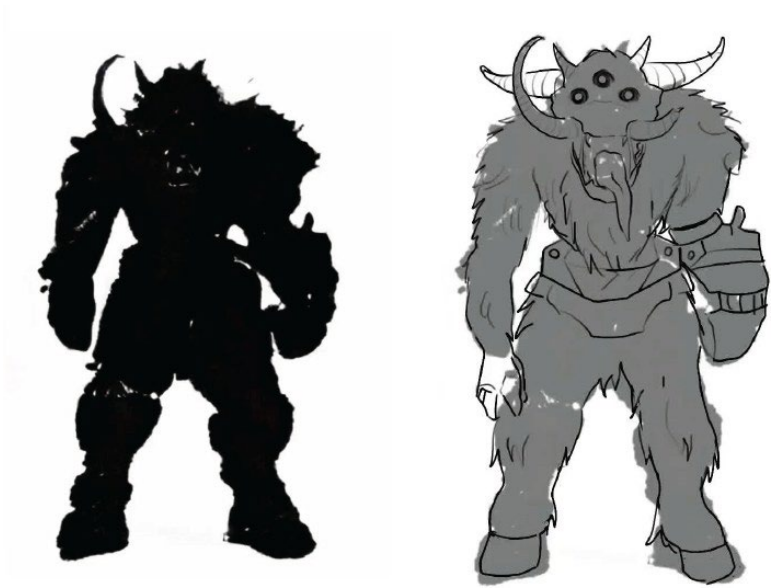

Figure A.24: Concept development from silhouette - Novice

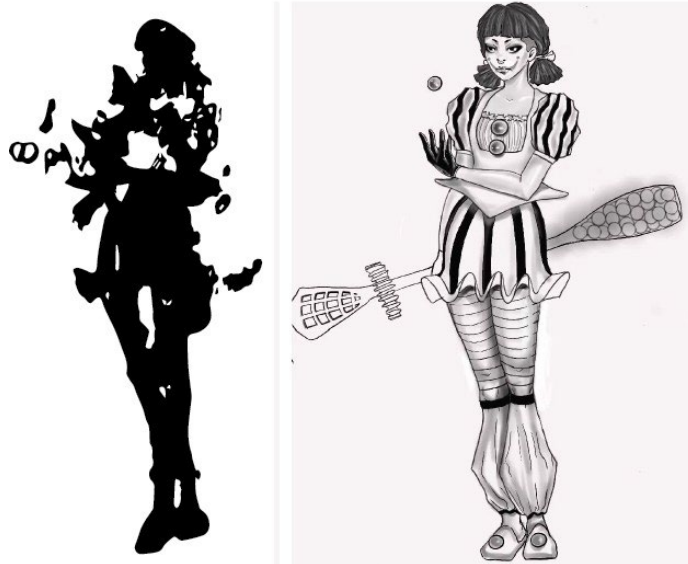

Figure A.25: Concept development from silhouette - Intermediate

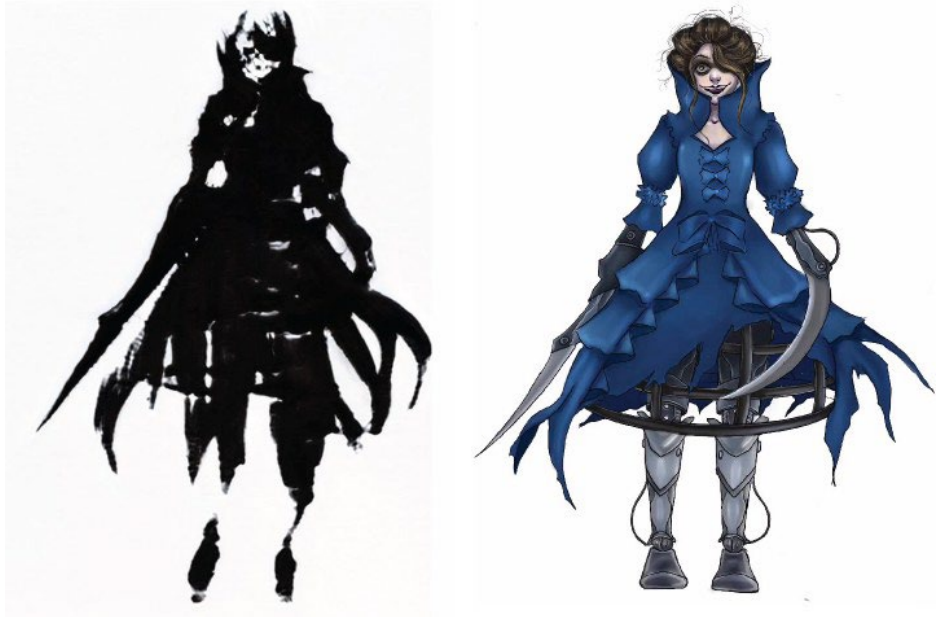

Figure A.26: Concept development from silhouette - Intermediate

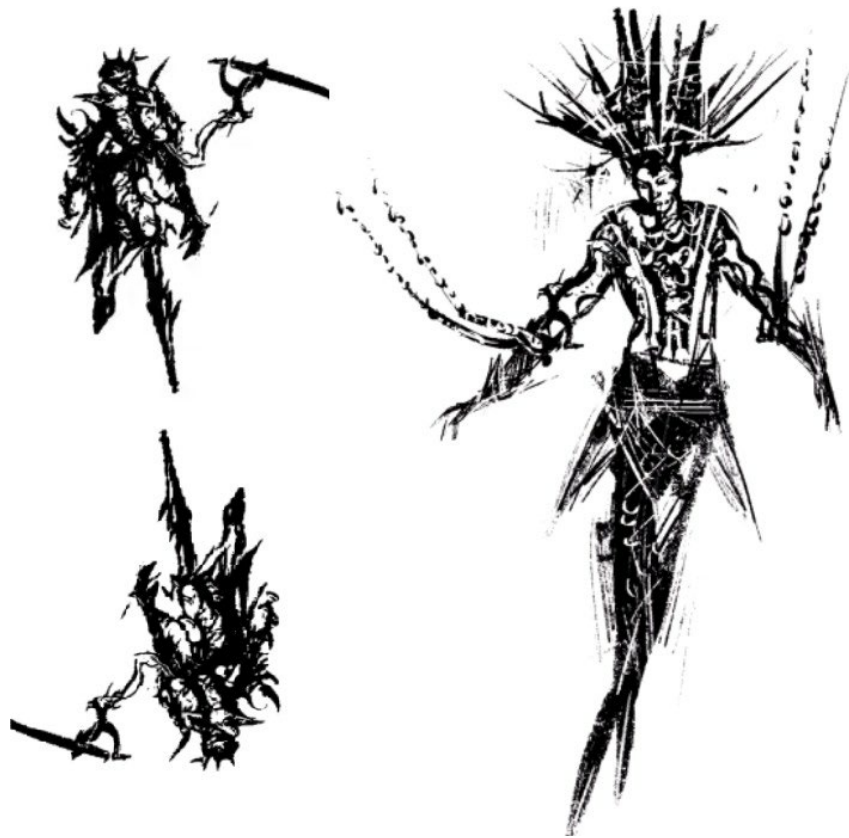

Figure A.27: Concept development from silhouette - Expert